\documentclass[journal]{IEEEtran}

\usepackage{url}
\usepackage{amsmath}
\makeatletter
\newif\if@restonecol
\makeatother

\usepackage[ruled,linesnumbered]{algorithm2e}
\usepackage{multirow}
\usepackage{array}
\usepackage{graphicx}
\usepackage{color}
\usepackage{cite}
\usepackage{bbding}

\begin{document}

\title{Dual-Refinement: Joint Label and Feature Refinement for Unsupervised Domain Adaptive Person Re-Identification}

\author{Yongxing Dai,~\IEEEmembership{Student Member,~IEEE,}
        Jun Liu,~\IEEEmembership{Member,~IEEE,}
        Yan Bai,~\IEEEmembership{Student Member,~IEEE,}
        
        Zekun Tong,
        and Ling-Yu Duan,~\IEEEmembership{Member,~IEEE}
\thanks{Y. Dai and Y. Bai are with the National Engineering Laboratory for Video Technology,
Peking University, Beijing 100871, China (e-mail: yongxing dai@pku.edu.cn; yanbai@pku.edu.cn).}
\thanks{J. Liu is with the Information Systems Technology and Design Pillar, Singapore University of Technology and Design, Singapore 487372 (e-mail: jun\_liu@sutd.edu.sg)}
\thanks{Z. Tong is with the Department of Industrial Systems Engineering and Management, National University of Singapore, Singapore 117576 (e-mail: zekuntong@u.nus.edu)}
\thanks{L.-Y. Duan is with the National Engineering Laboratory for Video Technology, Peking University, Beijing 100871, China, and also with the Peng Cheng
Laboratory, Shenzhen 518055, China (e-mail: lingyu@pku.edu.cn).}
}

\maketitle

\begin{abstract}
Unsupervised domain adaptive (UDA) person re-identification (re-ID) is a challenging task due to the missing of labels for the target domain data. To handle this problem, some recent works adopt clustering algorithms to off-line generate pseudo labels, which can then be used as the supervision signal for on-line feature learning in the target domain.
However, the off-line generated labels often contain lots of noise that significantly hinders the discriminability of the on-line learned features, and thus limits the final UDA re-ID performance. 
To this end, we propose a novel approach, called Dual-Refinement, that jointly refines pseudo labels at the off-line clustering phase and features at the on-line training phase, to alternatively boost the label purity and feature discriminability in the target domain for more reliable re-ID.
Specifically, at the off-line phase, a new hierarchical clustering scheme is proposed, which selects representative prototypes for every coarse cluster.
Thus, labels can be effectively refined by using the inherent hierarchical information of person images.
Besides, at the on-line phase, we propose an instant memory spread-out (IM-spread-out) regularization, that takes advantage of the proposed instant memory bank to store sample features of the entire dataset and enable spread-out feature learning over the entire training data instantly.
Our Dual-Refinement method reduces 
the influence of noisy labels and refines the learned features within the alternative training process. 
Experiments demonstrate that our method outperforms the state-of-the-art methods by a large margin.  
\end{abstract}


\IEEEpeerreviewmaketitle

\section{Introduction}

\IEEEPARstart{P}{erson} re-identification (re-ID) aims at identifying the same person's images as the query in a gallery database across disjoint cameras.
Due to the great value in practical applications concerning public security and surveillance, person re-ID has attracted booming attention in the research community \cite{zheng2016person, ye2020deep}. 
Most of the existing works for person re-ID focus on fully supervised scenarios \cite{wang2018learning,wei2018glad,sun2019learning, luo2019strong, chen2019abd, 9075365, 9117031}, where they have obtained superior performance on some benchmarks as the model's training and testing are conducted in the same domain. However, their performances often drop dramatically when
models trained on the labeled source domain are directly applied to the unlabeled target domain owing to the domain gap. To handle the domain gap issue in the cross-domain re-ID scenario, recently, plenty of unsupervised domain adaptation~(UDA)  approaches \cite{yang2019leveraging,yang2020part} have been proposed. 
Many UDA re-ID methods adopt clustering algorithms \cite{song2020unsupervised,fu2019self, zhang2019self, yang2019asymmetric, ge2020mutual} to generate pseudo labels for the unlabeled target domain data, which can then be used as the supervision signal for training. 

Concretely, in clustering-based UDA methods \cite{song2020unsupervised,fu2019self, zhang2019self, yang2019asymmetric, ge2020mutual}, the model is often pre-trained with the labeled source domain data via a fully supervised manner. Then the model is fine-tuned using unlabeled target domain data in an alternative training manner including the off-line pseudo label generation phase and on-line feature learning phase.
Specifically, at the off-line phase, pseudo labels are generated by performing clustering on the features of the target domain samples, which are extracted with the trained model. At the on-line phase, the model is trained under the supervision of the pseudo labels generated from the off-line phase. The off-line label generation and on-line model training are conducted alternatively and iteratively over the whole learning process.
However, the pseudo labels generated at the off-line phase often contain noise, that directly affects the on-line feature learning performance \cite{ge2020mutual}.
Meanwhile, the discriminability of the on-line learned features, in turn, 
affects the off-line pseudo label generation at the next epoch. 
Thus, we need to alleviate the influence of label noises at both off-line and on-line phases to improve the performance.
Therefore, in this paper, we propose a novel approach called Dual-Refinement to jointly refine the off-line pseudo label generation and also the on-line feature learning under the noisy pseudo labels, during an alternative training process.

\begin{figure}[t]
  \centering
\includegraphics[width=\linewidth]{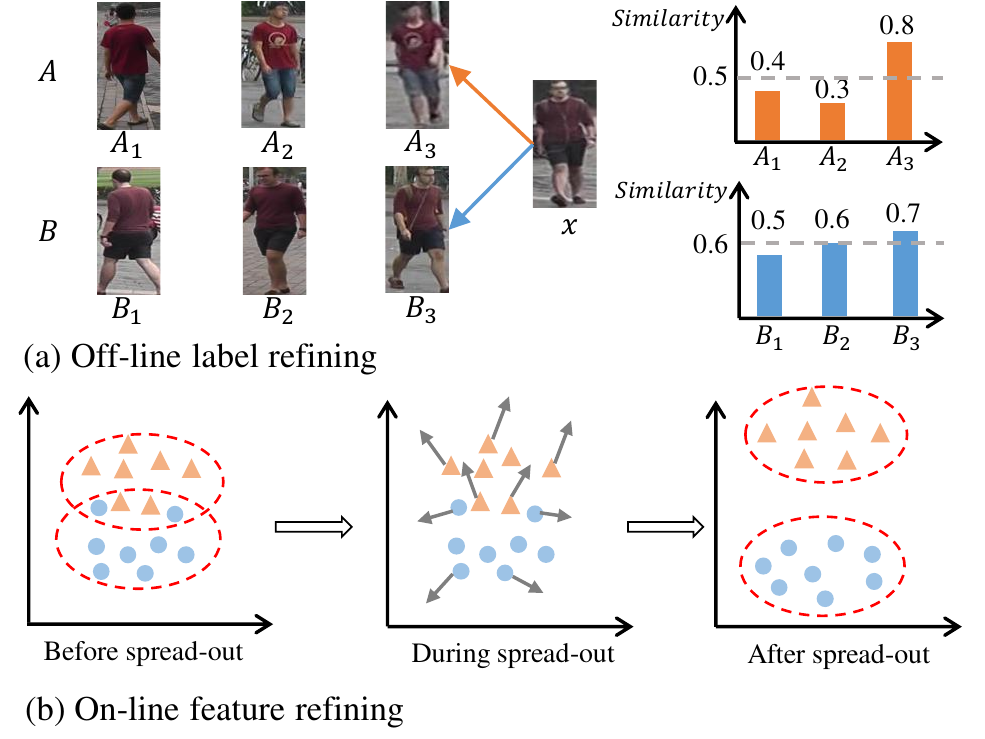}
\vspace{-0.4cm}
  \caption{
  (a) In our proposed hierarchical clustering method, cluster A and B are obtained by coarse clustering first. Then the cluster centers obtained with further fine clustering are selected as the representative prototypes of each coarse cluster. Here $A_{1}$, $A_{2}$, $A_{3}$ and $B_{1}$, $B_{2}$, $B_{3}$ are representative prototypes of clusters $A$ and $B$, respectively.
  Similarity histograms denote the similarity between the sample $x$ and each prototype. 
  Though $x$ seems to be more similar to $A_{3}$ than $B_{3}$, the average similarity between $x$ and cluster $B$'s prototypes ($B_{1}, B_{2}, B_{3}$) is 0.6, which is higher than the average similarity between $x$ and ($A_{1}, A_{2}, A_{3}$). By considering the global characteristics based on the representative prototypes, we can exploit more inherent and robust similarities.
  (b) Two-dimensional visualization of feature space. 
  Points and triangles denote two different clusters. Dashed ovals represent class decision boundaries. Arrows represent spreading out the features. By enforcing the spread-out property in one-line feature learning, the effect of noisy pseudo labels can also be alleviated.}
  \vspace{-0.5cm}
  \label{fig:intro-fig1}
\end{figure}

To refine the off-line pseudo labels, we design a novel hierarchical clustering scheme at the off-line phase.
Existing clustering-based UDA methods \cite{song2020unsupervised,fu2019self, zhang2019self} generally 
perform clustering based on the local similarities among the samples \cite{ester1996density} to assign pseudo labels, \textit{i.e.,} the sample's nearest neighborhoods are more likely to be grouped into the same cluster. 
However, such sample neighborhoods tend to ignore the global and inherent characteristics of each cluster, due to the high intra-cluster variance caused by different poses or viewpoints of the person. 
As shown in Fig. \ref{fig:intro-fig1} (a), when only considering the local similarities, the sample $x$ can be easily wrongly grouped into $A$, because $x$ and $A_{3}$ share the same pose and viewpoint and they are very similar.
As a result, the off-line pseudo label assignment based on such coarse clustering often brings about label noises.

To refine the noisy pseudo labels,
we propose to consider the global characteristics of every coarse cluster by selecting several representative prototypes within every coarse cluster. Specifically, we propose a new hierarchical clustering method, in which we perform fine clustering after the coarse clustering. Moreover, the fine sub-cluster centers serve as the representative prototypes for every coarse cluster. 
Compared with local similarities, the average similarity between the sample and these representative prototypes is more powerful in capturing each person's global and inherent information.
Thus it provides a more robust criterion for the off-line pseudo label assignment, and mitigates the label noise issue. 
As shown in Fig. \ref{fig:intro-fig1} (a), when considering the global characteristics within every cluster, the sample $x$ should be grouped into the cluster $B$, because the average similarity between $x$ and cluster $B$'s prototypes ($B_{1}, B_{2}, B_{3}$) is 0.6, which is larger than 0.5 achieved by using $A$'s prototypes. 
Thus by considering the global characteristics within each coarse cluster, we can assign more reliable pseudo labels.

At the on-line phase, to refine the feature learning and alleviate the effects of noisy pseudo labels,
we propose an instant memory spread-out (IM-spread-out) regularization scheme in our Dual-Refinement method. 
Since the pseudo labels can still contain noise, directly using such noisy pseudo labels to supervise the metric learning (classification and triplet loss) can limit the on-line feature learning performance. 
As shown in Fig. \ref{fig:intro-fig1} (b), the label noises originate from the off-line clustering that aims at discovering class decision boundaries \cite{caron2018deep,huang2019unsupervised}, and the noisy samples tend to be located near the decision boundary. Thus, these noisy samples confuse the on-line feature learning and limit the discriminability of the learned features. 
To pull noisy samples away from the decision boundaries and boost the discriminability of features, we propose an IM-spread-out regularization scheme during on-line feature learning, as shown in Fig.~\ref{fig:intro-fig1}~(b). 
In this paper, we define the spread-out property as separating those hard noisy samples further away from the cluster boundaries, where the hard noisy samples always locate near the cluster boundaries.
The spread-out property will not break the inherent characteristics of those reliable samples, because the reliable samples are inherently compact in a cluster, which thus is still robust during our IM-spread-out regularization.

To effectively capture the global distribution, we enforce the spread-out property on the whole training dataset.
Specifically, we consider every sample in the target domain as an instance, and our IM-spread-out regularization scheme satisfies the positive-centered and spread-out properties. 
Moreover, the sample's $k-$nearest neighborhoods can be seen as the positives, and all the remaining of the entire dataset can be seen as the negatives.
In this paper, the positive-centered property means that the $k$-nearest neighborhood samples should concentrate on the anchor sample to ensure the intra-class compactness in the embedding space.
However, it is hard to enforce the spread-out constraints on the whole training data in the mini-batch training manner, since the mini-batch only captures the local data distribution.
A possible solution is to use a memory bank  \cite{wu2018improving,wu2018unsupervised} to store the features of all the samples. 
However, existing memory bank mechanisms~\cite{wu2018unsupervised,zhong2019invariance,wang2020cross} are updated with outdated features in a momentum updating manner \cite{wu2018unsupervised, zhong2019invariance} or just using an enqueue-and-dequeue manner \cite{wang2020cross} to store the incoming features for collecting sufficient hard negative pairs across multiple mini-batches, all of which are not updated with back-propagating gradients and can not capture the global characteristics of the dataset. Thus, they are not suitable for our on-line feature learning.
Therefore, to enable effective spread-out constraint on the whole training data for alleviating the effects of noisy labels at the on-line stage, we propose a new \textbf{instant memory bank} that can be instantly updated together with the encoder by back-propagating gradients.

Our instant memory bank memorizes the sample features when features are fed into the bank, and meanwhile the instant memory bank is updated together with the network instantly at each training iteration. This means our memory bank always
stores the instant features (rather than outdated features) of all the samples. Thus it effectively and efficiently captures the global distribution. 
Thanks to the instant memory bank, our proposed IM-spread-out regularization effectively alleviates the effects of noisy supervision signal during each training iteration of the on-line stage, and further boosts the on-line features' discriminability.
To the best of our knowledge, this is the first memory bank mechanism that is able to maintain the \textit{instant features of all the training samples} during network optimization iterations, which thus greatly facilitates our spread-out scheme for feature optimization based on the global distribution.
The on-line feature refinement and off-line pseudo label refinement are conducted in an alternative manner. Finally, the trained model can generalize well in the target domain.

The major contributions can be summarized as follows:
\begin{itemize}
\item We propose a novel approach, called Dual-Refinement, to alleviate the pseudo label noise in clustering-based UDA re-ID, including the off-line pseudo label refinement to assign more accurate labels and the on-line feature refinement to alleviate the effects of noisy supervision. 

\item We design a hierarchical clustering scheme to select representative prototypes for every coarse cluster, which captures the more global and inherent characteristics of each person, thus can refine the pseudo labels at the off-line phase.

\item We propose an IM-spread-out regularization scheme to alleviate the effects of pseudo label noises at the on-line phase. Thus it improves the feature discriminability in the target domain. Moreover, a novel instant memory bank is proposed to store instant features and thus can enforce the spread-out property on the whole target training dataset.

\item Extensive experiments have shown that our method outperforms state-of-the-art UDA approaches significantly.

\end{itemize}

\section{Related Work}

\subsection{Unsupervised Domain Adaptation} 
The existing general UDA methods fall into two main categories: closed set UDA \cite{tzeng2014deep, ganin2015unsupervised, long2015learning, long2016unsupervised, kang2019contrastive} and open set UDA \cite{panareda2017open, saito2018open}. In closed set UDA, both the target and source domain completely share the same classes. Most of the closed set UDA works \cite{tzeng2014deep, ganin2015unsupervised, long2015learning} try to learn the domain invariant features to generalize the class decision boundary well on the target domain. 
Long \textit{et al.} propose DAN \cite{long2015learning} and RTN \cite{long2016unsupervised} to minimize Maximum Mean Discrepancy (MMD) across different domains. 
In open set UDA \cite{panareda2017open,saito2018open}, the target and source domain dataset only share a part of classes.
Saito \textit{et al.} \cite{saito2018open} use adversarial training to align target samples with known source samples or recognize them as an unknown class. 
All the general UDA methods mentioned above assume that the source and target domain share the whole or partial classes under the image classification scenario, which are difficult to be directly applied to UDA person re-ID tasks.

\subsection{Unsupervised Domain Adaptation for Person re-ID} 
The unsupervised domain adaptation methods for person re-ID can be mainly categorized into two aspects, one is the GAN-based \cite{deng2018image, wei2018person, zhong2018generalizing}, and the other is the clustering-based \cite{fu2019self, zhang2019self, yang2019asymmetric, song2020unsupervised, ge2020mutual}. SPGAN \cite{deng2018image} and PTGAN \cite{wei2018person} use CycleGAN \cite{CycleGAN2017} to translate the style of the source domain to the target domain and conduct the feature learning with the source domain labels. HHL \cite{zhong2018generalizing} uses StarGAN \cite{choi2018stargan} to learn the features with camera invariance and domain connectedness. UDAP \cite{song2020unsupervised} first proposes the clustering-based UDA framework for re-ID. 
SSG \cite{fu2019self} and PCB-PAST \cite{zhang2019self} bring the information of both the global body and local parts into the clustering-based framework. 
Some clustering-based methods \cite{ge2020mutual, yang2019asymmetric, zhang2019self} are devoted to solving the pseudo label noise problem.
MMT \cite{ge2020mutual} proposes an on-line peer-teaching framework to refine the noisy pseudo labels, which uses the on-line reliable soft labels generated from the temporally average model of one network to supervise the training of another network. 
However, both MMT \cite{ge2020mutual} and ACT \cite{yang2019asymmetric} introduce other networks that will bring about extra noises, and they are not memory efficient.
Besides, PCB-PAST \cite{zhang2019self} proposes the ranking-based triplet loss to alleviate the influence of label noises in on-line metric learning but it only focuses on local data distribution based on mini-batches. 
There are also some other works like ECN \cite{zhong2019invariance} and ECN+GPP \cite{zhong2020learning}, which use the traditional memory bank \cite{xiao2017joint} to consider every sample as an instance to learn the invariant feature. However, these methods need additional images generated by GAN,
and the features stored in the traditional memory bank are outdated \cite{he2020momentum}.

Different from all the UDA re-ID methods mentioned above, 
we propose a novel Dual-Refinement method that is able to jointly refine the pseudo labels at the off-line stage and alleviate the effects of noisy labels at the on-line stage to refine the on-line features. 
Moreover, different from traditional memory mechanisms \cite{xiao2017joint,zhong2019invariance} that update the memory with out-dated features and thus may lead to inconsistencies in feature learning due to discrepancies between the memory updating and the encoder updating, 
in this paper, we also propose a novel instant memory bank, which is instantly updated with the encoder by back-propagating gradients, and thus effectively enforces the spread-out property on the whole training data and captures the global characteristics of the whole target domain.

\begin{figure*}[htp]
  \centering
  \includegraphics[width=\linewidth]{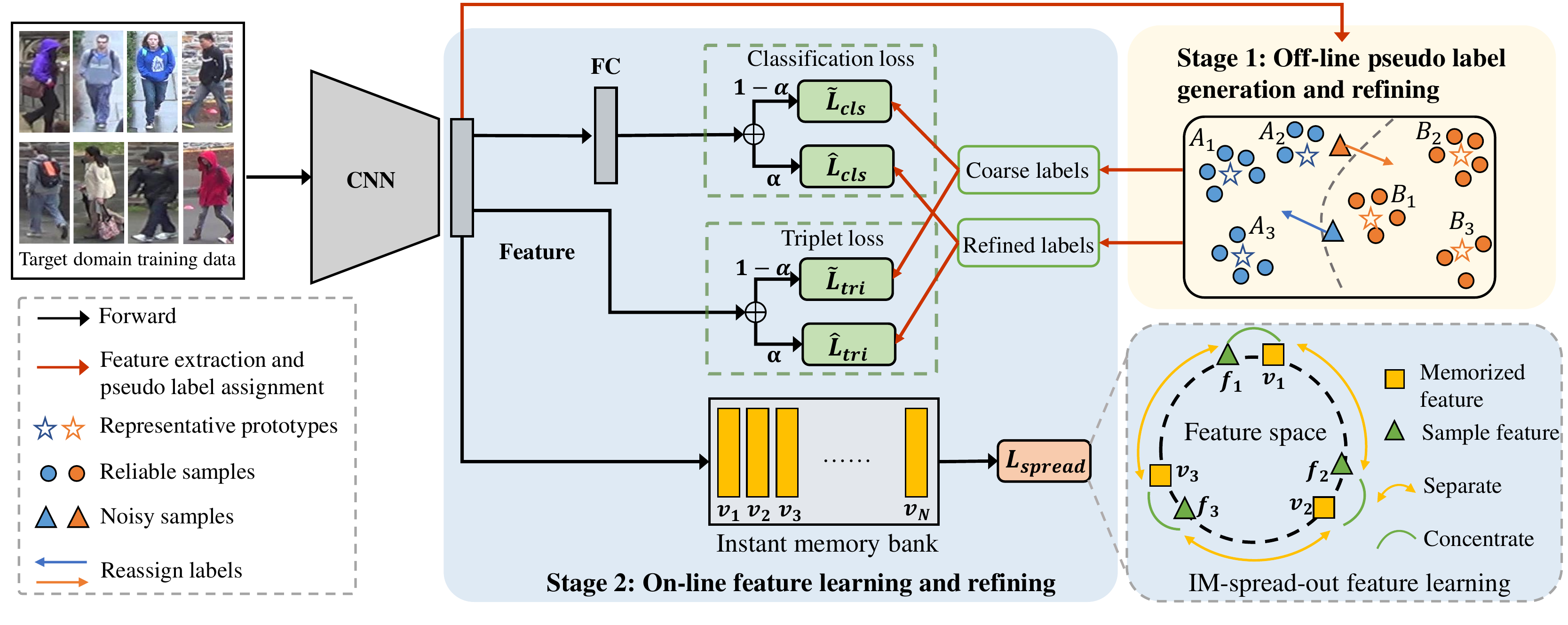}
  \caption{
  The framework of our method. 
  After initializing CNN by pre-training it on the labelled source domain, we train our network in the target domain in an alternative manner including two stages. 
  At the beginning of every training epoch, we conduct the off-line stage, where we use the trained model to extract all the sample features and then perform the hierarchical clustering on these features to assign pseudo labels.
  Then, we conduct the on-line stage, where we 
  use pseudo labels generated from the off-line stage to fine-tune the model with the classification loss and triplet loss, together with a label-free IM-spread-out regularization.
  These two stages are performed alternatively and iteratively in the target domain.
  The instant memory bank is used to store instant sample features. 
  The spread-out feature learning aims to separate different sample features (\textit{e.g.,} $f_{1}, f_{2}, f_{3}$) and concentrate the sample feature with its corresponding memory (\textit{e.g.,} $f_{1}, v_{1}$) in the feature space. Best viewed in color.
  }
  \vspace{-0.5cm}
  \label{fig:pipeline}
\end{figure*}

\subsection{Unsupervised Person re-ID}
Unsupervisedd person re-ID methods \cite{lin2019bottom,zeng2020hierarchical,wang2020unsupervised} does not use any labeled source data and they only use the model pretrained on ImageNet instead of other source re-ID data.
BUC~\cite{lin2019bottom} proposes a bottom-up clustering approach to jointly optimize the network and assign pseudo labels. 
In BUC, each image is first initialized as a cluster. Then they merge pairs of clusters into one based on their distance, \textit{i.e.,} the minimum distance criterion between the pair of clusters. However, BUC suffers from the quality of pseudo labels. To improve the quality of pseudo labels, HCT~\cite{zeng2020hierarchical} corrects false pseudo labels by evaluating on the model performance and cluster merging step. However, the hierarchical clustering mechanisms in BUC and HCT are both bottom-up, which can only consider the inter-sample and inter-cluster similarities while neglecting the intra-cluster diversity. Different form them, our proposed hierarchical clustering mechanism is top-down. We first obtain coarse clusters and then select robust representative prototypes in individual coarse clusters, which can capture the global and inherent characteristics of the target training data. In MMCL~\cite{wang2020unsupervised}, samples' neighborhood similarity and cycle consistency are utilized to improve the quality of predicted multi-class labels but both of them do not encourage the well-clustered goal. Thus, MMCL cannot ensure that the feature space is highly concentrated among class centers. Different from MMCL, our proposed method aims to learn well-clustered feature space to ensure the discriminability in the target domain.

\subsection{Learning with Noisy Labels}

Existing works on learning with noisy labels can be categorized into four main groups. The methods in the first category focus on learning a transition matrix \cite{menon2015learning, goldberger2016training, patrini2017making, xia2019anchor}. 
However, it is hard to estimate the noise label transition for UDA re-ID because the classes in the target domain are unknown. 
The second category \cite{ghosh2015making, zhang2018generalized,szegedy2016rethinking} is to design the loss functions robust to noise labels, but they bring about extra constraints like the mean absolute loss in GCE \cite{zhang2018generalized}. 
The third category \cite{lee2018cleannet, han2018co, jiang2018mentornet} is to utilize additional networks to refine the noisy labels. Co-teaching \cite{han2018co} uses two networks in a co-trained manner.
These methods need other networks and complicated sample selection strategies. The last category \cite{reed2014training,tanaka2018joint,han2019deep} learns on noisy labels in a self-training manner. 
Han \textit{et al.} \cite{han2019deep} propose the SMP method to select several class prototypes for each category to correct the noisy labels in fully supervised learning. Specifically, SMP randomly samples images in each class to calculate the similarity matrix for prototypes' selection. However, SMP only exploits the neighborhood similarity of a small portion of training data, which can not capture the global distribution within each class.
Different from the aforementioned methods in a supervised image classification scenario, we propose a hierarchical clustering scheme to select robust prototypes to represent the global characteristics in each pseudo class. As a result, these robust class prototypes can handle the label noise issue in UDA re-ID.

\subsection{Feature Embedding with Spread-out Property} 
Feature embedding learning with the spread-out property has improved the performance in deep local feature learning \cite{zhang2017learning}, unsupervised embedding learning \cite{ye2019unsupervised}, and face recognition \cite{liu2018learning, zhao2019regularface,duan2019uniformface}. Zhang \textit{et al.} \cite{zhang2017learning} propose a Global Orthogonal Regularization to fully utilize the feature space by making the negative pairs close to orthogonal. 
Ye \textit{et al.} \cite{ye2019unsupervised} use a siamese network to learn data augmentation invariant and instance spread-out features under the instance-wise
supervision.
These works \cite{zhang2017learning,ye2019unsupervised} guarantee the spread-out property of features within a mini-batch. Specifically, in \cite{ye2019unsupervised}, only the augmented sample can be seen as the positive and the remaining samples within a mini-batch are the negatives, where the limited number of positive and the negative samples may lead to less discriminative feature learning. 
Other works in face recognition \cite{liu2018learning, zhao2019regularface, duan2019uniformface} use a term regularized on the classifier weights to make class centers spread-out in the holistic feature space. However, they are under the supervision of the ground truth. 
Unlike the above works, our IM-spread-out regularization is used to alleviate the influence of noisy labels on on-line metric learning for UDA re-ID. Besides, we use an instant memory bank to enforce the spread-out property on the entire training data instead of the mini-batch, where the positive samples are the top-k similar ones selected from the memory bank and all the remaining of the whole training data are the negative ones. Thanks to the instant memory bank, the samples' diversity can further boost the spread-out feature learning.

\section{Approach}

In UDA re-ID, we are given a labeled source domain dataset and an unlabeled target domain dataset. In the source domain, the dataset $ D_{s}=\{(x_{i}^{s},y_{i}^{s})|_{i=1}^{N_{s}} \}$ contains $ N_{s} $ person images and each image $ x_{i}^{s} $ corresponds to an identity label $ y_{i}^{s} $. 
In the target domain, dataset $ D_{t}= \{ x_{i}^{t} |_{i=1}^{N_{t}} \} $ contains $ N_{t} $ unlabeled person images. Our goal is to use both labeled source data and unlabeled target data to learn discriminative image representations in the target domain.

\subsection{Overview of Framework}

As shown in Fig. \ref{fig:pipeline}, the framework of our method contains two stages including the off-line pseudo label refinement stage and the on-line feature learning stage. 
The network (CNN) is initialized by pre-training on source domain data, following a similar method in \cite{luo2019strong}.
CNN is a deep feature encoder $ F(\cdot |\theta) $ parameterized with $ \theta $, and can encode the person image into a $d$-dimensional feature. 

At the off-line stage, we propose a hierarchical clustering scheme for the target domain features $\{f_{i}^{t}|_{i=1}^{N_{t}}\}$ extracted by the network (CNN) trained in the last epoch, where $ f_{i}^{t}=F(x_{i}^{t}|\theta) $. By clustering, we assign samples in the same cluster with the same pseudo label, and then each target domain sample gets two kinds of pseudo labels, including noisy label $ \widetilde{y}_{i}^{t}\in \{ 1,2,...,L^{t} \} $ and refined label $ \widehat{y}_{i}^{t}\in \{ 1,2,...,L^{t} \} $, where $ L^{t} $ is the number of the unique labels. The off-line pseudo labels are used for the on-line feature learning.

At the on-line stage, we use samples with noisy labels $ \widetilde{D}_{t}=\{(x_{i}^{t},\widetilde{y}_{i}^{t})|_{i=1}^{N_{t}} \} $ to train with the classification loss $\widetilde{\mathcal L}_{cls}$ and triplet loss $\widetilde{\mathcal L}_{tri}$, and use samples with refined labels $ \widehat{D}_{t}=\{ (x_{i}^{t},\widehat{y}_{i}^{t})|_{i=1}^{N_{t}} \} $ to train with the loss $\widehat{\mathcal L}_{cls}$ and $\widehat{\mathcal L}_{tri}$.
FC is a $ L^{t} $ dimensional fully connected layer followed by softmax function, which is denoted as the identity classifier $ \phi $.
To alleviate the influence of pseudo label noises in the on-line supervised metric learning,
we propose a label-free IM-spread-out regularization scheme $\mathcal L_{spread}$ to train together with the classification loss and triplet loss \cite{hermans2017defense}. 
Specifically, we propose a novel instant memory bank  parameterized by $ V=\{ v_{i}|_{i=1}^{N_{t}} \} $ to store all the samples' features instantly. Thus, the instant memory bank can enforce the spread-out property on the whole target training dataset instead of the mini-batch, which can capture the global characteristics of the target domain distribution. 
We first conduct the off-line stage at the beginning of every epoch and conduct the on-line stage during every epoch, both of which are conducted in an alternative and iterative manner.
For simplicity, we omit the superscript $ t $ for the target domain data in the following sections.

Below, we first introduce the general clustering-based UDA procedure. We then introduce our Dual-Refinement method that can optimize both stages in the cluster-based UDA procedure, \textit{i.e.,} jointly refine off-line pseudo labels and on-line features, and thus improve the overall UDA performance.

\subsection{General Clustering-based UDA Procedure }
\label{sec:general-cluster}
Existing clustering-based UDA methods \cite{song2020unsupervised,ge2020mutual,zhang2019self} for re-ID usually pre-train the backbone network with classification loss and triplet loss on labeled source dataset, and then use the pre-trained model to initialize the model for training the target dataset $ D= \{ x_{i} |_{i=1}^{N} \} $. We follow a similar method in \cite{luo2019strong} to pre-train the backbone model with the labeled source domain, and then perform the training procedure on the target domain, which contains two stages: (1) Off-line assigning pseudo labels based on clustering at the beginning of every training epoch. (2) Utilizing target domain data with pseudo labels to train the network with metric learning loss on-line during every training epoch. These two stages are conducted alternatively and iteratively in the training process.

\subsubsection{Off-line assigning pseudo labels} Following the existing clustering based UDA methods \cite{song2020unsupervised,zhang2019self,fu2019self}, we first extract features $ \{ f_{1}, f_{2}, ..., f_{N} \} $ of all $ N $ images in the target domain using the CNN trained from last epoch, and then calculate their pair-wise similarity $ d_{S}(i,j) $ between samples $i$ and $j$ by:
\begin{equation}
d_{S}(i,j)=\left\{  
\begin{array}{lr}
e^{-\left \| f_{i}-f_{j}\right \|_{2}}, & if \ j\in R^{*}(i,k)\\
0, & otherwise \ .
\end{array}
\right.
\label{eq:kencoding}
\end{equation}
$ R^{*}(i,k) $ is the $k$-reciprocal nearest neighbor set of sample $x_{i}$ introduced in \cite{zhong2017re}. We then use the pair-wise similarity to calculate the Jaccard distance $d_{J}(i,j)$ by:
\begin{equation}
    d_{J}(i,j)=1-\frac{\sum_{k=1}^{N}min(d_{S}(i,k),d_{S}(j,k))}
   {\sum_{k=1}^{N}max(d_{S}(i,k),d_{S}(j,k))} \ ,
   \label{eq:jaccard}
\end{equation}
where $N$ is the number of target training samples. We use such distance metric $d_{J}(i,j)$ to perform DBSCAN clustering \cite{ester1996density} on target domain and obtain $L$ clusters. 
We consider each cluster as a unique class and assign the same pseudo label for the samples belonging to the same cluster. Thus, the target domain images with their pseudo labels are represented as $ \widetilde{D}=\{(x_{i},\widetilde{y}_{i})|_{i=1}^{N} \} $, where $ \widetilde{y}_{i}\in \{ 1,2,...,L \}$.

\subsubsection{On-line model training with metric learning losses} In this step, 
we use the target dataset $ \widetilde{D}=\{(x_{i},\widetilde{y}_{i})|_{i=1}^{N} \} $ labeled by pseudo labels $\{\widetilde{y}_{i}\}$ to train the network with  classification loss $\mathcal L_{cls} $ and triplet loss \cite{hermans2017defense} $\mathcal L_{tri} $, which are foumulated as follows: 
\begin{equation}
    \widetilde{\mathcal L}_{cls}=\frac{1}{N}\sum_{i=1}^{N}\mathcal L_{ce}(\phi (f_{i}),\widetilde{y}_{i}) \ ,
\label{eq:noisy-cls}
\end{equation}

\begin{equation}
\widetilde{\mathcal L}_{tri} = \frac{1}{N}\sum_{i=1}^{N}max(0,\Delta+\left \| f_{i}-f_{i}^{+}\right \|_{2}-\left \| f_{i}-f_{i}^{-}\right \|_{2}) \ ,
\label{eq:noisy-tri}
\end{equation}
where $\phi$ denotes the classification layer FC and $\mathcal L_{ce}$ denotes the cross-entropy loss. $f_{i}^{+}$ and $f_{i}^{-}$ are the features of the hardest positive and negative of the sample $x_{i}$ through batch hard mining \cite{hermans2017defense} under the supervision of pseudo labels $ \{ \widetilde{y}_{i} \} $, and $ \Delta $ is the margin. We denote this general UDA re-ID method as the \textbf{baseline} in this paper and fine-tune the network with an overall loss $\mathcal L_{baseline} $, that is obtained by combining the loss $\widetilde{\mathcal L}_{cls}$ and $\widetilde{\mathcal L}_{tri}$:
\begin{equation}
    \mathcal L_{baseline}=\widetilde{\mathcal L}_{cls}+\widetilde{\mathcal L}_{tri} \ .
\end{equation}

\subsection{Off-line Pseudo Label Refinement}
\label{sec:off-line}
At the off-line stage, we design a hierarchical clustering guided label refinement strategy. 
The refining process contains two stages of hierarchical clustering from coarse (\textit{i.e.,} DBSCAN) to fine (\textit{i.e.,} K-means). 
For coarse clustering, DBSCAN is used to find the global class boundaries and filter some easy noisy samples (\textit{i.e.,} outliers) because DBSCAN has a notion of noise and is robust to outliers. However, only using coarse clustering may neglect inta-cluster diversity and cause many hard samples to locate near class boundaries.
Therefore, we design a fine clustering using K-means to select representative prototypes within each coarse cluster, which is inspired by recent works in semi-supervised learning~\cite{kuo2020featmatch} and label noise reduction~\cite{sharma2020noiserank}. 
K-means is a centroid-based clustering method which groups data into K clusters and obtains K cluster centers by minimizing within-cluster variances. As a result, the  K cluster centers of each coarse cluster can serve as the prototypes, which are robust to the intra-cluster variances and can capture the global characteristics of each coarse cluster. Because of these representative prototypes, we can refine the noisy labels by utilizing the intra-cluster diversity. Specifically, fine pseudo labels are assigned based on the average similarity between the noisy sample and prototypes in each coarse cluster.

Following the pseudo label assignment in Section \ref{sec:general-cluster}, we can obtain $L$ coarse clusters and the target sample set of the class $l$ is denoted as $ D_{l}=\{ (x_{i},\widetilde{y}_{i})|\forall \widetilde{y}_{i}=l  \} $. 
We extract features of each class to constitute the feature set $ F_{l}=\{ (f_{i},\widetilde{y}_{i})|\forall \widetilde{y}_{i}=l  \} $ and then perform K-means clustering \cite{lloyd1982least} on every coarse class feature set $ F_{l} $, 
\textit{i.e,} the coarse cluster $l$ will be splited into R sub-clusters. After such fine clustering, we pick the center of every sub-cluster as a prototype. Thus, we can obtain $R$ representative prototypes $ \{ c_{l,1},c_{l,2},...,c_{l,R} \} $ for every coarse cluster $l$. By considering all the prototypes, we can get more global and inherent characteristics of every coarse cluster.

As shown in Fig. \ref{fig:intro-fig1} (a), the global characteristics within each coarse cluster can help assign more reliable pseudo labels.
Thus, we define the refined similarity score of sample $ x_{i} $ belonging to class $ l $ as $ s_{i,l} $, which is calculated by
\begin{equation}
s_{i,l}=\frac{1}{R}\sum_{r=1}^{R}f_{i}^{T}c_{l,r} \ .
\label{eq:refined-similarity}
\end{equation}

The feature $f_{i}$ and the prototype are all L2-normalized. Similarity score $ s_{i,l} $ can be viewed as the average similarity between the sample $ x_{i} $ and the representative prototypes of class $l$. Instead of assigning the coarse label for all samples within a cluster, our refined similarity score $ s_{i,l} $ can provide a more reliable similarity between every sample and coarse clusters. With this similarity, we can reassign more reliable labels for target data by
\begin{equation}
    \widehat{y}_{i}=\underset{l}{arg \ max}\ s_{i,l}\  \ ,l\in \{1,2,...,L\} \ ,
    \label{eq:reliable-labels}
\end{equation}
where $ \widehat{y}_{i} $ is the refined pseudo label of sample $ x_{i} $. Now every target sample $x_{i}$ has two kinds of pseudo labels including coarse noisy label $ \widetilde{y}_{i} $ and refined label $ \widehat{y}_{i} $.

\subsection{On-line Feature Refinement}
\subsubsection{Metric learning with pseudo labels} 
We can use the refined labels to optimize the network with metric learning losses including $\widehat{\mathcal L}_{cls}$ and $ \widehat{\mathcal L}_{tri} $, where $\widehat{\mathcal L}_{cls}$ and $ \widehat{\mathcal L}_{tri} $ are obtained by replacing the coarse pseudo labels 
$ \{\widetilde{y}_{i} \} $ with our refined labels $ \{\widehat{y}_{i} \}$ in Eq. (\ref{eq:noisy-cls}) (\ref{eq:noisy-tri}).

We combine metric losses under the supervision of both noisy pseudo labels and refined pseudo labels by:
\begin{equation}
\left\{\begin{matrix}
\mathcal L_{cls}=(1-\alpha )\widetilde{\mathcal L}_{cls}+\alpha\widehat{\mathcal L}_{cls} \\ 
\mathcal L_{tri}=(1-\alpha )\widetilde{\mathcal L}_{tri}+\alpha\widehat{\mathcal L}_{tri}
\end{matrix}\right. \ .
\label{eq:metric}
\end{equation}

We use $\alpha \in [0,1]$ to control the weights of reliable and noisy pseudo labels on both classification loss and triplet loss. 
When $\alpha=0$, it means we only use the supervision of the coarse pseudo labels. When $\alpha=1$, it means we only use the supervision of the fine pseudo labels. 
If $0<\alpha<1$, both the coarse pseudo labels and the refined pseudo labels are used to prevent the network from learning with the over-confident hard noisy samples, which can be seen as a calibration on the metric losses. We use Eq. (\ref{eq:metric}) to include any of the above situations to provide the scalability of our method. 
Coarse pseudo labels contain the inter-cluster information and refined pseudo labels contain the intra-cluster information, which can complement each other to alleviate the effects of learning with pseudo labels during training. More analysis on the parameter $\alpha$ can be seen at Section \ref{sec:param-alpha}.

\subsubsection{IM-spread-out regularization with instant memory bank} 
Although we have designed the off-line pseudo label refining strategy, it is not possible to eliminate all the label noises.
To alleviate the effects of noisy labels on feature learning, we propose label-free regularization that aims to spread out the features in the whole feature space and pull the samples assigned with false labels out of the same class. 
Though spread-out properties with mini-batches have shown the effectiveness in recent works \cite{duan2019uniformface, ye2019unsupervised},
in our task, to capture the whole characteristics of the target domain for more effective spread-out, we enforce the spread-out property on the entire target training dataset, instead of the mini-batch.

To capture the whole characteristics of the target domain, one possible solution is to use a memory bank  \cite{wu2018improving,wu2018unsupervised} to store features of all the samples. However, traditional memory bank mechanisms \cite{xiao2017joint, wu2018improving,zhong2019invariance} can only memorize \textit{outdated features} \cite{he2020momentum}, 
because each entry in a traditional memory bank is updated only once in each epoch, while the network is contiguously updated at every iteration. 
This means there are discrepancies between the memory updating and the encoder updating, leading to inconsistencies in on-line feature learning \cite{he2020momentum}.
Hence, such memory bank methods are not suitable for handling our on-line feature refinement problem. To this end, here we propose a new instant memory bank that is able to store the \textit{instant features} of all the samples, \textit{i.e.,} all the entries in the bank are updated together with the network instantly at every iteration.

As shown in Fig. \ref{fig:pipeline}, we propose the instant memory bank $V$, where each entry in 
$\{ v_{i}|_{i=1}^{N} \}$ is a $d$-dimensional vector, \textit{i.e.,} $v_{i} \in R^{d}$. 
We use $v_{i}$ to approximate the feature $f_{i} \in R^{d}$ and thus the memory bank $V$ can memorize the approximated features of the entire dataset.
For simplicity, we use the L2-normalized feature via $f_{i}\leftarrow f_{i}/\left \| f_{i}\right \|_{2}$ and $v_{i}\leftarrow v_{i}/\left \| v_{i}\right \|_{2}$. To make the memory entries approximate the sample features more accurately, the similarity between entry $v_{i}$ and feature $f_{i}$ should be as large as possible, \textit{i.e.,} $f_{i}^{T}v_{i}$ is close to 1. 

The spread-out property means the feature $f_{i}$ of every sample $x_{i}$ in the entire training dataset should be dissimilar with each other, \textit{i.e.,} $f_{i}^{T}f_{j}$ should be close to -1 when $j\neq i$.
To further improve the discrimination of the feature learning, the feature should satisfy not only the spread-out property but also the positive-centered property. We assume that the $k$-nearest neighborhoods of the sample $x_{i}$ in memory belong to the same class, where we denote the index set of $k$-nearest neighborhoods as $\mathcal{K}_{i}$. We can consider all the samples in $\mathcal{K}_{i}$ along with $x_{i}$ itself as the positives of $x_{i}$ (\textit{i.e.,} $\mathcal{K}_{i}\leftarrow \mathcal{K}_{i}\cup \{i\}$) and the samples not in $\mathcal{K}_{i}$ as the negatives of $x_{i}$. To concentrate the positives and separate the negatives far away, our IM-spread-out regularization can be formulated as:
\begin{equation}
\mathcal L_{spread}=\frac{1}{N}\sum_{i=1}^{N}log[1+\sum_{k\in \mathcal{K}_{i}}\sum_{n=1 \atop n\notin \mathcal{K}_{i}}^{N}exp(f_{i}^{T}v_{n}-f_{i}^{T}v_{k}+m)] \ ,
\label{eq:spreadout}
\end{equation}
where $m$ is the margin to spread the negatives. 
The calculation of Eq. (\ref{eq:spreadout}) with the instant memory bank can be easily implemented by normal matrix operations, which is training efficient.
With the batch gradient scheme, we calculate the derivatives of the instant memory bank entry $v_{j}$ by:
\begin{equation}
\frac{\partial \mathcal L_{spread}}{\partial v_{j}}
=\frac{1}{N}\sum_{i=1}^{N}T_{j}/ [ 1+\sum_{k\in \mathcal{K}_{i}}\sum_{n=1 \atop n\notin \mathcal{K}_{i}}^{N}exp(f_{i}^{T}v_{n}-f_{i}^{T}v_{k}+m) ] \ ,
\label{eq:gradient1}
\end{equation}
\begin{equation}
T_{j}=\left\{  
\begin{array}{lr}  
\sum_{n=1\atop n\notin \mathcal{K}_{i}}^{N}exp(f_{i}^{T}v_{n}-f_{i}^{T}v_{j}+m)\cdot (-f_{i}), &if \  j\in \mathcal{K}_{i} \\
\sum_{k\in \mathcal{K}_{i}}exp(f_{i}^{T}v_{j}-f_{i}^{T}v_{k}+m)\cdot f_{i}, &if \  j\notin \mathcal{K}_{i}  
\end{array}  
\right.
\label{eq:gradient2}      
\end{equation}

All the entries $\{v_{j}\}$ in our instant memory bank are updated instantly by Eq. (\ref{eq:gradient1}) (\ref{eq:gradient2}) together with the network at every training iteration, which can be formulated as follows:

\begin{equation}
v_{j}=v_{j}-\eta\frac{\partial \mathcal L_{spread}}{\partial v_{j}},\ \ 
v_{j}=\frac{v_{j}}{||v_{j}||_{2}},
\label{eq:update-instant-memory}
\end{equation}
where $\eta$ represents the learning rate of the network. With Eq.~(\ref{eq:update-instant-memory}), we update together with the L2-normalized instant features of all the training samples, which is able to effectively capture the characteristics of the whole target domain distribution in real time.

By performing spread-out regularization with the instant features of all the training samples using our instant memory bank,
the effects of the noisy pseudo labels on on-line feature learning can thus be alleviated, and the features' discriminability can be further boosted.
Note that our spread loss can also be seen as a variant of the circle loss \cite{sun2020circle}, yet it is significantly different from the original circle loss, as the circle loss needs to be trained with mini-batch in a fully supervised manner.

\subsubsection{Overall loss} 
By combining Eq. (\ref{eq:metric}) under the supervision of pseudo labels and the label-free regularization Eq. (\ref{eq:spreadout}) together,
the overall objective loss is formulated as:
\begin{equation}
\begin{split}
&\mathcal L_{joint}=\mathcal L_{cls}+\mathcal L_{tri}+\mu \mathcal L_{spread}\\
&=(1-\alpha )(\widetilde{\mathcal L}_{cls}+\widetilde{\mathcal L}_{tri})+\alpha(\widehat{\mathcal L}_{cls}+\widehat{\mathcal L}_{tri})+\mu \mathcal L_{spread} \ ,
\end{split}
\label{eq:overall-loss}
\end{equation}
where $\alpha$ and $\mu$ are the parameters to balance the losses. 
Our proposed off-line pseudo label refinement and on-line feature refinement are conducted alternatively and iteratively over the whole learning process. 
The details about the overall training procedure can be seen in Algorithm \ref{alg:algorithm1}.

\begin{algorithm}[tp]
\small
\label{alg:algorithm1}
    \caption{Alternative Training Procedure of Our Dual-Refinement Method}
    \label{alg:algorithm1}
    \KwIn{Labeled source dataset $D_{s}$; Unlabeled target dataset $D$ with $N$ images; Feature encoder $F$ pretrained on ImageNet; Identity classifier $\phi$; Instant memory bank $V$; 
    Maximum training epoch $max\_epoch$; Maximum training iteration $max\_iter$}
    \KwOut{Optimized feature encoder $F$\ for target domain;}
    Pretain feature encoder $F$ on the labeled source dataset $D_{s}$\ with classification loss and triplet loss;
    
    Use the source-pretrained encoder $F$ to extract all target samples' features $\{f_{i}|_{i=1}^{N}\}$, and initialize the instant memory bank entry $v_{i}$ as the corresponding $f_{i}$.
    
    \For{$epoch=1$ to $max\_epoch$}
    {
        \textit{// Off-line pseudo label generation and refinement}
        
        Extract features of the target dataset $D$ using $F$ and calculate Jaccard distance $d_{J}(i,j)$ by Eq. (\ref{eq:kencoding}) (\ref{eq:jaccard});
        
        Perform DBSCAN clustering on $D$ with $d_{J}$ and assign the coarse pseudo labels $\widetilde{D}=\{(x_{i},\widetilde{y}_{i})|_{i=1}^{N} \}$;
        
        Perform fine clustering and assign refined pseudo labels  by Eq. (\ref{eq:refined-similarity}) (\ref{eq:reliable-labels}) to obtain the refined target dataset $\widehat{D}=\{(x_{i},\widehat{y}_{i})|_{i=1}^{N} \}$;
        
        \textit{// On-line feature learning and refinement}
        
        \For{$iter=1$ to $max\_iter$}
        {
        Sample $(x_{i},\widetilde{y}_{i},\widehat{y}_{1})$ from $\widetilde{D} \cup \widehat{D}$ \;
        Update the feature encoder $F$, classifier $\phi$ and instant memory bank $V$ by computing the gradients of the overall loss (Eq. (\ref{eq:overall-loss})) with back-propagation;
        }
    }
    \textbf{return} feature encoder $F$;
\end{algorithm}

\section{Experiments}
\subsection{Datasets and Evaluation Protocol}

We conduct experiments on three large-scale person re-ID datasets namely Market1501 \cite{zheng2015scalable}, DukeMTMC-ReID \cite{zheng2017unlabeled} and MSMT17 \cite{wei2018person}.
The mean average precision (mAP) and Cumulative Matching Characteristic (CMC) curve \cite{gray2007evaluating} are used as the evaluation metrics.
Specially, we use the rank-1 accuracy (R1), rank-5 accuracy (R5) and rank-10 accuracy (R10) in CMC.
There is no post-processing like re-ranking \cite{zhong2017re} applied at the testing stage.

Market1501 \cite{zheng2015scalable} contains 32,668 labeled images of 1,501 identities captured from 6 different camera views. All the person images are detected by a Deformable Part Model. The training set consists of 12,936 images of 751 identities and the testing set consists of 19,732 images of 705 identities.

DukeMTMC-ReID \cite{zheng2017unlabeled} contains 36,411 labelled images of 1,404 identities which are captured from 8 different camera views. It is also a subset from the DukeMTMC dataset. The training set consists of 16,522 images of 702 identities for training. The testing set consists of 2,228 query images of 702 identities and 17,661 gallery images.

\begin{table*}[htp]
\caption{Comparisons between the proposed method and  state-of-the-art unsupervised domain adaptation methods for person re-ID.
The best results are highlighted with bold and the second best results are highlighted with underline. '-' indicates the results not reported}
\small
\label{tab:SOTA}
\begin{center}
\begin{tabular}{l|c|p{1.1cm}<{\centering}p{1.1cm}<{\centering}p{1.1cm}<{\centering}p{1.1cm}<{\centering}|p{1.1cm}<{\centering}p{1.1cm}<{\centering}p{1.1cm}<{\centering}p{1.1cm}<{\centering}}
\hline
\multirow{2}{*}{Methods} & \multirow{2}{*}{Reference} & \multicolumn{4}{c|}{DukeMTMC-ReID $\to$ Market1501} & \multicolumn{4}{c}{Market1501 $\to$ DukeMTMC-ReID} \\ \cline{3-10} 
                         &                            & mAP      & R1       & R5       & R10     & mAP      & R1       & R5      & R10     \\ \hline \hline
LOMO \cite{liao2015person}                     & CVPR 2015                  & 8.0      & 27.2     & 41.6     & 49.1    & 4.8      & 12.3     & 21.3    & 26.6    \\
BOW \cite{zheng2015scalable}                      & ICCV 2015                  & 14.8     & 35.8     & 52.4     & 60.3    & 8.3      & 17.1     & 28.8    & 34.9    \\ \hline
UMDL \cite{peng2016unsupervised}                     & CVPR 2016                  & 12.4     & 34.5     & 52.6     & 59.6    & 7.3      & 18.5     & 31.4    & 37.6    \\
PTGAN \cite{wei2018person}                    & CVPR 2018                  & -        & 38.6     & -        & 66.1    & -        & 27.4     & -       & 50.7    \\
PUL \cite{fan2018unsupervised}                     & TOMM 2018                  & 20.5     & 45.5     & 60.7     & 66.7    & 16.4     & 30.0     & 43.4    & 48.5    \\
SPGAN \cite{deng2018image}                   & CVPR 2018                  & 22.8     & 51.5     & 70.1     & 76.8    & 22.3     & 41.1     & 56.6    & 63.0    \\
ATNet \cite{liu2019adaptive}                   & CVPR 2019                  & 25.6     & 55.7     & 73.2     & 79.4    & 24.9     & 45.1     & 59.5    & 64.2    \\
TJ-AIDL \cite{wang2018transferable}                 & CVPR 2018                  & 26.5     & 58.2     & 74.8     & 81.1    & 23.0     & 44.3     & 59.6    & 65.0    \\
SPGAN+LMP \cite{deng2018image}                & CVPR 2018                  & 26.7     & 57.7     & 75.8     & 82.4    & 26.2     & 46.4     & 62.3    & 68.0    \\
CamStyle \cite{zhong2019camstyle}                & TIP 2019                   & 27.4     & 58.8     & 78.2     & 84.3    & 25.1     & 48.4     & 62.5    & 68.9    \\
HHL \cite{zhong2018generalizing}                     & ECCV 2018                  & 31.4     & 62.2     & 78.8     & 84.0    & 27.2     & 46.9     & 61.0    & 66.7    \\
ECN \cite{zhong2019invariance}                     & CVPR 2019                  & 43.0     & 75.1     & 87.6     & 91.6    & 40.4     & 63.3     & 75.8    & 80.4    \\
PDA-Net \cite{li2019cross}                 & ICCV 2019                  & 47.6     & 75.2     & 86.3     & 90.2    & 45.1     & 63.2     & 77.0    & 82.5    \\
UDAP \cite{song2020unsupervised}                    & PR 2020                    & 53.7     & 75.8     & 89.5     & 93.2    & 49.0     & 68.4     & 80.1    & 83.5    \\
PCB-PAST \cite{zhang2019self}                & ICCV 2019                  & 54.6     & 78.4     & -        & -       & 54.3     & 72.4     & -       & -       \\
SSG \cite{fu2019self}                     & ICCV 2019                  & 58.3     & 80.0     & 90.0     & 92.4    & 53.4     & 73.0     & 80.6    & 83.2    \\
MMCL \cite{wang2020unsupervised}                 & CVPR 2020                 & 60.4     & 84.4     & 92.8     & 95.0    & 51.4     & 72.4     & 82.9    & 85.0    \\
ACT \cite{yang2019asymmetric}                 & AAAI 2020                 & 60.6     & 80.5     & -     & -    & 54.5     & 72.4     & -    & -    \\
ECN-GPP \cite{zhong2020learning}                 & TPAMI 2020                 & 63.8     & 84.1     & 92.8     & 95.4    & 54.4     & 74.0     & 83.7    & \underline{87.4}    \\
AD-Cluster \cite{zhai_adcluster}              & CVPR 2020                  & 68.3     & 86.7     & 94.4     & 96.5    & 54.1     & 72.6     & 82.5    & 85.5    \\
MMT \cite{ge2020mutual}                     & ICLR 2020                  & \underline{71.2}     & \underline{87.7}     & \underline{94.9}     & \underline{96.9}    & \underline{65.1}     & \underline{78.0}     & \underline{88.8}    & \textbf{92.5}    \\ \hline
Dual-Refinement                     & This paper                 & \textbf{78.0}     & \textbf{90.9}     & \textbf{96.4}     & \textbf{97.7}    & \textbf{67.7}     & \textbf{82.1}     & \textbf{90.1}    & \textbf{92.5}    \\ \hline \hline
\multirow{2}{*}{Methods} & \multirow{2}{*}{Reference} & \multicolumn{4}{c|}{DukeMTMC-ReID $\to$ MSMT17}     & \multicolumn{4}{c}{Market1501 $\to$ MSMT17}   \\ \cline{3-10} 
                         &                            & mAP      & R1       & R5       & R10     & mAP      & R1       & R5      & R10     \\ \hline \hline
ECN \cite{zhong2019invariance}                     & CVPR 2019                  & 10.2     & 30.2     & 41.5     & 46.8    & 8.5      & 25.3     & 36.3    & 42.1    \\
SSG \cite{fu2019self}                     & ICCV 2019                  & 13.3     & 32.2     & -        & 51.2    & 13.2     & 31.6     & -       & 49.6    \\
ECN-GPP \cite{zhong2020learning}                 & TPAMI 2020                 & 16.0     & 42.5     & 55.9     & 61.5    & 15.2     & 40.4     & 53.1    & 58.7    \\
MMCL \cite{wang2020unsupervised}                 & CVPR 2020                 & 16.2     & 43.6     & 54.3     & 58.9    & 15.1     & 40.8     & 51.8    & 56.7    \\
MMT \cite{ge2020mutual}                     & ICLR 2020                  & \underline{23.3}     & \underline{50.1}     & \underline{63.9}     & \underline{69.8}    & \underline{22.9}     & \underline{49.2}     & \underline{63.1}    & \underline{68.8}    \\ \hline
Dual-Refinement                     & This paper                 & \textbf{26.9}     & \textbf{55.0}     & \textbf{68.4}     & \textbf{73.2}    & \textbf{25.1}     & \textbf{53.3}     & \textbf{66.1}    & \textbf{71.5}   \\ \hline
\end{tabular}
\end{center}
\end{table*}

MSMT17 \cite{wei2018person} contains 126,441 images of 4,101 identities. The images are captured by 15 cameras  during 4 days, in which 12 cameras are outdoor and 3 cameras are indoor. The bounding box of every person is detected by Faster RCNN.  The training set contains 32,621 images of 1,041 identities, and the testing set contains 11,659 query images and 82,161 gallery images. MSMT17 is now the large-scale dataset that poses greater challenge to cross-domain person re-ID compared to the other two datasets mentioned above.

\subsection{Implementation Details}

We utilized ResNet50 \cite{he2016deep} pre-trained on ImageNet \cite{deng2009imagenet} as the backbone network. We add a batch normalization (BN) layer followed by ReLU after the global average pooling (GAP) layer. The stride size of the last residual layer is set as 1. The identity classifier layer is a fully connected layer (FC) followed by softmax function. We resize the image size to $256\times 128$. For data augmentation, we perform random cropping, random flipping, and random erasing \cite{zhong2017random}. The margin $\Delta$ in Eq.~\ref{eq:noisy-tri} is set as 0.3, and the margin $m$ in Eq.~\ref{eq:spreadout} is set as 0.35. If not specified, we set the parameter $\alpha=0.5$ and $\mu=0.1$ to balance the joint loss in Eq (\ref{eq:overall-loss}). We use the Adam \cite{kingma2014adam} optimizer with weight decay $5\times 10^{-4}$ and momentum $0.9$ to train the network. 
The learning rate in the pre-training stage follows the warmup learning strategy where the learning rate linearly increases from $3.5\times10^{-5}$ to $3.5\times10^{-4}$ during the first 10 epochs.
The learning rate is divided by 10 at the 40th epoch and 70th epoch, respectively, in a total of 80 epochs. 
We set the batch size as 64 in all our experiments.
When training on target data, the learning rate $\eta$ is initialized as $3.5\times10^{-4}$ and divided by 10 at the 20th epoch in a total of 40 epochs. During testing, we extract the L2-normalized feature after the BN layer and use Euclidean distance to measure the similarity between the query and gallery images in the testing set. 
Our model is implemented on PyTorch \cite{paszke2019pytorch} platform and trained with 4 NVIDIA TITAN XP GPUs.

\begin{table*}[htp]
\caption{Ablation studies on supervised, direct transfer and variants combined with baseline. LR means off-line pseudo label refinement with hierarchical clustering. IM-SP means on-line feature refinement with the  IM-spread-out regularization in Eq. (\ref{eq:spreadout}). Our method (Baseline with both LR and IM-SP) is comparable to the fully supervised methods.}
\small
\label{tab:ablation}
\begin{tabular}{l|p{1.1cm}<{\centering}p{1.1cm}<{\centering}p{1.1cm}<{\centering}p{1.1cm}<{\centering}|p{1.1cm}<{\centering}p{1.1cm}<{\centering}p{1.1cm}<{\centering}p{1.1cm}<{\centering}}
\hline
\multicolumn{1}{c|}{\multirow{2}{*}{Methods}} & \multicolumn{4}{c|}{DukeMTMC-ReID$\to$Market1501} & \multicolumn{4}{c}{Market1501$\to$DukeMTMC-ReID} \\ \cline{2-9} 
\multicolumn{1}{c|}{}                         & mAP      & R1       & R5       & R10     & mAP      & R1       & R5       & R10     \\ \hline \hline
Fully Supervised (upper bound)                             & 81.2     & 93.1     & 97.7     & 98.6    & 70.3     & 84.2     & 91.7     & 93.9    \\
Direct Transfer (lower bound)                              & 28.6     & 58.0     & 73.7     & 79.8    & 27.6     & 44.5     & 60.6     & 66.1    \\ \hline
Baseline                                      & 67.9         & 85.7         & 94.3         & 96.3        & 56.4         & 72.5         & 84.5         & 88.2        \\
Baseline with only LR                                   & 74.4         & 88.7         & 95.1         & 97.1         & 65.5          & 80.0         & 89.8         & 92.7        \\
Baseline with only IM-SP                                   & 75.5         & 89.0         & 95.8         & 97.5         & 66.3         & 80.5         & 89.6         & 92.4         \\
Baseline with both LR and IM-SP                                & \textbf{78.0}         & \textbf{90.9}         & \textbf{96.4}         & \textbf{97.7}        & \textbf{67.7}         & \textbf{82.1}         & \textbf{90.1}         & \textbf{92.5}        \\
 \hline
\end{tabular}
\end{table*}

\subsection{Comparisons with State-of-the-Arts}

\subsubsection{Results on Market1501 and DukeMTMC-ReID} In Table \ref{tab:SOTA}, we compare our method with state-of-the-art methods.
GAN-based methods include PTGAN \cite{wei2018person}, SPGAN \cite{deng2018image}, ATNet \cite{liu2019adaptive}, CamStyle \cite{zhong2019camstyle}, HHL \cite{zhong2018generalizing} and PDA-Net \cite{li2019cross}; UDAP \cite{song2020unsupervised}, PCB-PAST \cite{zhang2019self}, SSG \cite{fu2019self}, ACT \cite{yang2019asymmetric} and MMT \cite{ge2020mutual} are based on clustering; AD-Cluster \cite{zhai_adcluster} combines GAN and clustering; ECN \cite{zhong2019invariance}, ECN-GPP \cite{zhong2020learning} and MMCL \cite{wang2020unsupervised} use memory bank. 
Our method achieves the performance of 78.0\% on mAP and 90.9\% on rank-1 accuracy when DukeMTMC-ReID$\to$Market1501. Our method outperforms state-of-the-art GAN-based method AD-Cluster \cite{zhai_adcluster} by 9.7\% on mAP and 4.2\% on rank-1 accuracy. Compared with the best clustering-based method MMT \cite{ge2020mutual}, our method is more than 6.8\% on mAP and 3.2\% on rank-1 accuracy when DukeMTMC-ReID$\to$Market1501. Besides, our method outperforms the best memory-bank-based method ECN-GPP \cite{zhong2020learning} by 14.2\% on mAP and 6.8\% on rank-1 accuracy when DukeMTMC-ReID$\to$Market1501. When using Market1501 as the source dataset and DukeMTMC-ReID as the target dataset, we get the performance of 67.7\% on mAP and 82.1\% on rank-1 accuracy, which is 13.6\% and 9.5\% higher than AD-Cluster \cite{zhai_adcluster}.
Compared with state-of-the-art UDA methods, our method improves the performance by a large margin. It should be noted that our method uses a single model in training and does not use any other images generated by GAN. 
However, MMT \cite{ge2020mutual} uses dual
networks for target domain training, where the amount of parameters is more than twice that of our method. For ECN-GPP \cite{zhong2020learning}, its performance heavily depends on the quality of extra augmented images produced by GAN.

\subsubsection{Results on MSMT17}
Our method still outperforms state-of-the-art methods on this challenging dataset by a large margin. When considering DukeMTM-ReID as the source dataset, our method achieves the performance of 26.9\% on mAP and 55.0\% on rank-1 accuracy, which is 3.6\% and 4.9\% higher than state-of-the-art MMT \cite{ge2020mutual}. When using Market1501 as the source dataset, we get 25.1\% performance on mAP and 53.3\% on rank-1 accuracy, which surpasses MMT \cite{ge2020mutual} by 2.2\% and 4.1\%. The improvement of the performance in such a challenging dataset has strongly demonstrated the effectiveness of our method.

\subsection{Ablation Study}

In this section, there are extensive ablation studies aiming at evaluating the effectiveness of different components of our method. 

\subsubsection{Comparisons between supervised learning, direct transfer, and baseline} In Table \ref{tab:ablation}, 
we compare the performance on fully supervised learning, direct transfer method and the baseline method mentioned in Section \ref{sec:general-cluster}. 
The fully supervised method can be seen as the upper bound for UDA of re-ID, and they use the ground-truth of the target domain to train with classification loss and triplet loss suggested in \cite{luo2019strong}. 
However when this source-trained model directly apply to the target domain, there is a huge performance gap because of the domain bias. 
When the model is trained on DukeMTMC-ReID and directly tested on Market1501, mAP drops from 70.3\% to 28.6\%. Furthermore, rank-1 accuracy drops from 84.2\% to 58.0\%. The baseline method uses the coarse clustering to assign the noisy pseudo labels and trains the model with classification loss and triplet loss, which is mentioned in Section \ref{sec:general-cluster}. The baseline method improves the performance by a larger margin compared with the direct transfer method. When transferring from DukeMTMC-ReID to Market1501, mAP and rank-1 accuracy of the baseline are 39.3\% and 27.7\% higher than the direct transfer method. Due to the huge domain gap, there is still a large margin in performance when comparing the baseline and the fully supervised method.
It should be noted that the performance of our baseline is comparable to state-of-the-art methods \cite{zhong2020learning, zhai_adcluster} trained with extra images generated by GANs, as shown in Table \ref{tab:SOTA}.

\subsubsection{Effectiveness of the off-line pseudo label refinement} In Table \ref{tab:ablation}, we evaluate the effectiveness of the off-line pseudo label refinement, denoted as LR in Table \ref{tab:ablation}. 
Baseline with only LR means that we only conduct the off-line refinement to generate pseudo labels to supervised the on-line metric learning.
When testing on DukeMTMC-ReID, Baseline with only LR outperforms the baseline method by 9.1\% on mAP and 7.5\% on rank-1 accuracy. It has shown that the hierarchical clustering guided pseudo label refinement plays an important role in off-line pseudo label refinement, which will promote the on-line discriminative feature learning.

\subsubsection{Effectiveness of the on-line feature refinement} In Table \ref{tab:ablation} we denote the IM-spread-out regularization as IM-SP. 
When only considering the on-line feature refinement, \textit{i.e.,} Baseline with only IM-SP, it is 7.6\% higher on mAP and 3.3\% higher on rank-1 accuracy than Baseline, which is tested on DukeMTMC-ReID$\to$Market1501.
It shows that the on-line feature refinement can also improve the performance based without the off-line refinement. 
From the performance of our method, \textit{i.e.,} Baseline with both LR and IM-SP, we can conclude that both the off-line label and on-line feature refinement are indispensable in our Dual-Refinement method.

\subsubsection{Comparisons between the  IM-spread-out regularization and its variants} In Table \ref{tab:memory-vs-proxy}, we compare different implementations and variants of the IM-spread-out regularization. 
The method SP+TM represents substituting the traditional memory bank \cite{xiao2017joint} for our instant memory bank, whose performance is 2.9\% lower than our method (SP+IM) on mAP when tested on Market1501. It shows that the out-dated features stored in memory bank can degenerate the performance. 
We also compare with the spread-out regularization training with mini-batch (SP+MB) and the invariance loss \cite{zhong2019invariance, zhong2020learning} equipped with traditional memory bank (IN+TM) in Table \ref{tab:memory-vs-proxy}. However, the performances of IN+TM evaluated on DukeMTMC-ReID and Market1501 are all worse than ours. From the above analysis, we can see that the invariance loss, training with mini-batch or the traditional memory bank can not learn the features as discriminative as our method.
We observe that the invariance loss is not able to learn more discriminative features, when compared to spread-out regularization. Training with mini-batch is not able to fully utilize the whole feature space, because the spread-out property is only enforced within a mini-batch. Due to the limited spread-out property, SP+MB is inferior to SP+TM.
No matter using our proposed spread-out regularization (SP) or using the invariance loss (IN), training with the instant memory bank outperforms training with the traditional memory bank. This shows the superiority of IM owing to its instant updating mechanism. However,
the traditional memory banks cannot update together with the encoder with back-propagation, which leads to  inconsistencies in on-line feature learning. As a result, SP+TM (IN+TM) is inferior to SP+IM (IN+IM).

\begin{table}[]
\caption{Analysis on our IM-spread-out regularization and its variants. SP: Spread-out regularization. IN: Invariance loss. IM: Instant memory bank. MB: Mini-batch. TM: Traditional memory bank.  }
\small
\label{tab:memory-vs-proxy}
\begin{tabular}{l|cc|cc}
\hline
\multirow{2}{*}{Method} & \multicolumn{2}{l|}{Duke$\to$Market1501} & \multicolumn{2}{l}{Market1501$\to$Duke} \\ \cline{2-5} 
& mAP & R1 & mAP & R1  \\ \hline \hline
Ours + SP + IM  & \textbf{78.0} & \textbf{90.9} & \textbf{67.7} & \textbf{82.1} \\
Ours + SP + TM  & 75.1 & 88.7 & 66.3 & 80.9  \\
Ours + SP + MB  & 73.2 & 88.4 & 62.6 & 77.2  \\ \hline
Ours + IN + IM &77.0 &89.8 &67.2 &80.5 \\ 
Ours + IN + TM  & 74.9  & 89.2  & 66.2 & 79.7              \\ 
Ours + IN + MB &74.0 &88.5 &61.9 &78.6 \\
\hline
\end{tabular}
\end{table}

\subsubsection{Analysis on the quality of pseudo labels} 
\begin{figure}[htp]
  \centering
  \includegraphics[width=\linewidth]{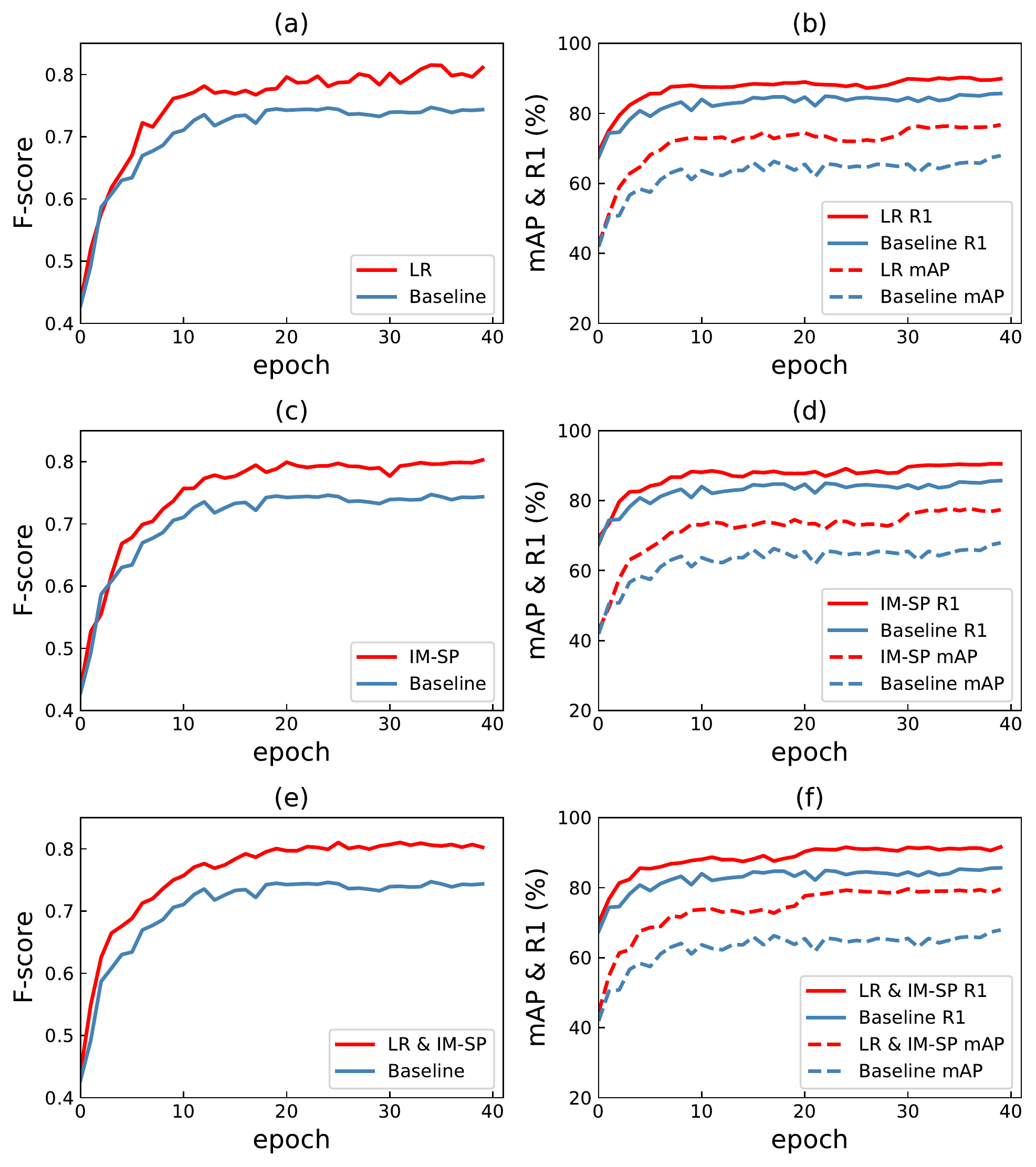}
 \caption{
 Evaluation on the effects of the quality of pseudo labels on DukeMTMC-ReID$\to$Market1501. (a) (c) (e) F-score evaluated on off-line pseudo label refinement (LR), on-line IM-spread-out feature refinement (IM-SP), and combination of both of them (LR \& IM-SP) respectively. (b) (d) (f) The performance comparing between baseline and LR, IM-SP, LR \& IM-SP respectively during training.
 }
  \label{fig:fscore}
\end{figure}

In Fig. \ref{fig:fscore}, we evaluate the effects of the pseudo labels' quality. We use \textit{Pairwise F-score} 
\cite{shi2018face}
to evaluate the quality of clustering, which is commonly used to evaluate the quality of Face Clustering \cite{shi2018face,yang2020learning} and the pseudo labels’ quality in UDA re-ID \cite{yang2019asymmetric,zhai_adcluster}.
Following the definition of \textit{Pairwise F-score} in \cite{shi2018face}, we define the ground truth ``positive pair'' and ``negative pair'' based on the criterion whether a pair of samples share the same person identity (obtained by the annotation of the training data). Specifically, we can obtain all $\frac{1}{2}(N-N_{outlier})\cdot (N-N_{outlier}-1)$ pairs in the target training data, where $N$ is the number of the whole target training data and $N_{outlier}$ is the number of the outliers detected by DBSCAN in coarse clustering. When using the pseudo labels and the true labels of all the target training samples, we can define the True Positive
Pairs (TP), False Positive Pairs (FP) and False Negative Pairs (FN). Precision is defined as the fraction of pairs that are labeled correctly together over the total number of pairs that share the same identity. Recall is defined as the fraction of pairs that are labeled correctly together over the total number of pairs that share the same pseudo label.
Thus, we can calculate \textit{Precision} and \textit{Recall} by:
\begin{equation}
    Precision=\frac{TP}{TP+FP} \ ,
\end{equation}

\begin{equation}
    Recall=\frac{TP}{TP+FN} \ .
\end{equation}

The \textit{F-score} is the harmonic mean of Precision and Recall and can be calculated by:
\begin{equation}
    \textit{F-score}=\frac{2\times Precision\times Recall}{Precision+Recall} \ .
\end{equation}

We evaluate the clustering quality (\textit{i.e., F-score}) and accuracy (\textit{i.e.,} mAP and R1) on off-line pseudo label refinement (LR), on-line IM-spread-out feature refinement (IM-SP), and combination of both of them (LR \& IM-SP) respectively in Fig. \ref{fig:fscore}. As shown in Fig. \ref{fig:fscore}, the pseudo label quality of LR, IM-SP, and LR \& IM-SP outperform the Baseline method by a large margin, where higher \textit{F-score} towards 1.0 implies better clustering quality and less noise in pseudo labels. Because of the increasing quality in pseudo labels, the performances (\textit{i.e.,} mAP and R1) also increase during training.
As shown in Fig. \ref{fig:fscore} (c) (d), our proposed IM-spread-out regularization scheme (IM-SP) can also improve pseudo label qualities and thus improve the target domain's discriminability. Specifically, IM-SP can separate those hard noisy samples further away from the class boundaries, which tend to be located near the decision boundary after the off-line clustering. 
By jointly considering to refine the off-line pseudo labels and alleviate the effects of the label noises on the on-line feature learning, our Dual-Refinement method (LR \& IM-SP) can improve the performance in the target domain.

\begin{table}[htp]
\caption{Analysis on different off-line clustering mechanisms. ``\XSolidBrush'' means not performing fine clustering, \textit{i.e.,} only using coarse clustering. ``\Checkmark'' means performing fine clustering after coarse clustering.}
\begin{tabular}{lc|cc|cc}
\hline
\multicolumn{2}{c|}{Clustering mechanism} & \multicolumn{2}{c|}{Duke$\to$Market1501} & \multicolumn{2}{c}{Market1501$\to$Duke} \\ \hline
\multicolumn{1}{c}{Coarse} & Fine & mAP & R1 & mAP & R1 \\ \hline \hline
DBSCAN &\XSolidBrush  &75.5  &89.0  &66.3  &80.5  \\
K-means-500 &\XSolidBrush  &71.3  &85.6  &59.3  &76.5  \\
K-means-700 &\XSolidBrush  &75.2  &88.1  &66.6  &80.4  \\
K-means-900 &\XSolidBrush  &74.1  &87.8  &66.5  &80.6  \\ \hline
DBSCAN &\Checkmark  &78.0  &90.9  &67.7  &82.1  \\
K-means-500 &\Checkmark  &71.5  &86.5  &60.5  &77.2  \\
K-means-700 &\Checkmark  &77.1  &89.5  &67.2  &81.2  \\
K-means-900 &\Checkmark  &77.0  &89.4  &67.5  &81.0  \\ \hline
\end{tabular}
\label{tab:compare-clustering}
\end{table}

\subsubsection{Comparison between different off-line clustering schemes}
As shown in Table \ref{tab:compare-clustering}, we can see that the performance is largely influenced by the number K if using K-means for coarse clustering. For example, if only performing coarse clustering, K-means-700 significantly outperforms K-means-500 by 7.3\% (mAP) and 3.9\% (R1) on Market1501$\to$Duke. We choose DBSCAN for coarse clustering in our methods instead of K-means because DBSCAN does not need to specify the number of coarse clusters in prior. 
Though the performances of K-means-900 are similar to those of DBSCAN, it is still hard to select an appropriate number K of clusters. No matter using DBSCAN or K-means for coarse clustering, our proposed hierarchical clustering mechanism can outperform those only performing coarse clustering. Take K-means-900 as an example, the methods using fine clustering can outperform those without fine clustering by 2.9\% (mAP), 1.6\% (R1) on Duke$\to$Market1501, and 1.0\% (mAP), 0.4\% (R1) on Market1501$\to$Duke.

\subsection{Further Analysis and Discussion}

\subsubsection{Comparisons with the traditional memory bank under different momentum hyper-parameters}
\begin{figure} [htp]
  \centering
  \includegraphics[width=\linewidth]{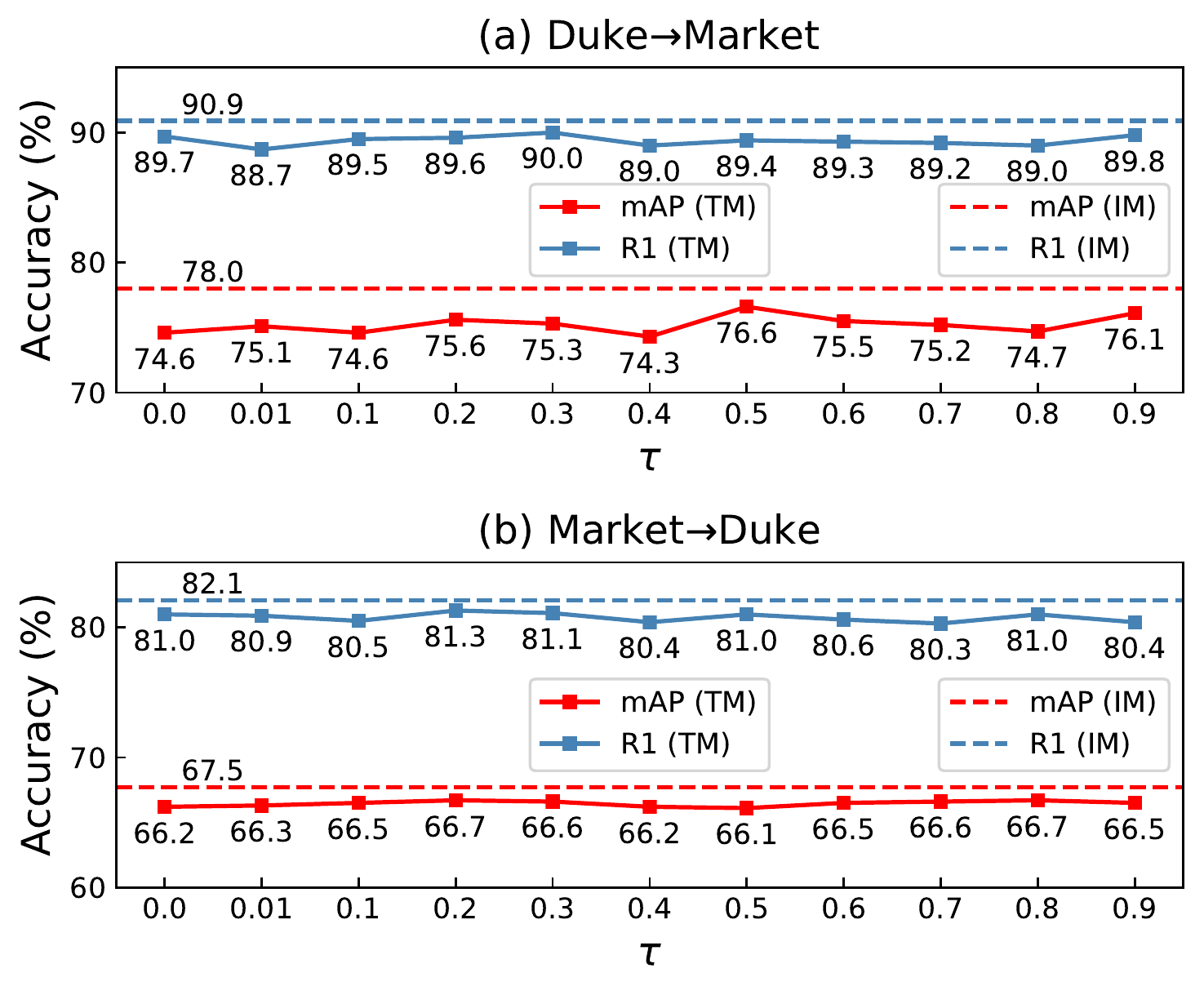}
  \caption{Comparisons between our proposed instant memory bank (IM) and the traditional memory bank (TM) with different momentum hyper-parameter~$\tau$.}
  \label{fig:param-momentum}
\end{figure}

The tradition memory bank is updated with a momentum manner as: $v_{i}\leftarrow \tau \cdot v_{i}+(1-\tau)\cdot f_{i}$, where $v_{i}$ is the entry of the memory bank, $f_{i}$ is the encoded sample feature, and $\tau$ is the momentum updating hyper-parameter. If not specified, we set $\tau$ as 0.01 in all experiments about traditional memory bank in our manuscript. As shown in Fig. (\ref{fig:param-momentum}), we tune the momentum hyper-parameter $\tau$ and compare their performances (TM) to the instant memory bank (IM). The number of $\tau$ is critical to the performance of the traditional memory bank, but our instant memory bank can be updated together with the network without tuning other hyper-parameters. Our instant memory bank shows its superiority to the traditional memory bank with different $\tau$.

\subsubsection{Computational cost comparisons}

As shown in Table \ref{tab:timecost}, we compare our Dual-Refinement method with the baseline and the state-of-the-art method MMT \cite{ge2020mutual} in the computional cost. The experiments are conducted when Market1501$\to$DukeMTMC-ReID. MMT uses two networks to train with each other, which is not memory efficient. Compared with MMT, our Dual-Refinement can achieve higher performance by costing less training time and GPU memory. Compared with the baseline method, our Dual-Refinement only introduces little extra GPU memory cost (about 908 MB) and little extra time cost (about 0.36 hours) because of the proposed instant memory bank. However, our Dual-Refinement outperforms the baseline method's rank-1 accuracy (R1) by a large margin. Based on the above analyses, our proposed Dual-Refinement is superior not only in the performance but also in the computational cost.

\begin{table}[tp]
\caption{Computional cost comparisions.}
\label{tab:timecost}
\begin{tabular}{l|ccc}
\hline
\multicolumn{1}{c|}{\multirow{2}{*}{Method}} & \multicolumn{3}{c}{Market1501 $\to$ DukeMTMC-ReID}               \\ \cline{2-4} 
\multicolumn{1}{c|}{}                        & R1 (\%) & Time (hours) & \multicolumn{1}{l}{GPU Memory (MB)} \\ \hline \hline
Baseline                                         & 72.5    & 3.17        & 8692                            \\
Dual-Refinement                                         & 82.1   & 3.53        & 9600                            \\
MMT                                          & 78.0     & 11.45       & 15068                           \\ \hline
\end{tabular}
\end{table}

\subsection{Parameter Analysis.} 
\label{sec:parameter-analysis}

In this section, we evaluate the influences of four hyper-parameters including the weight $\alpha$, the weight $\mu$ in Eq. (\ref{eq:overall-loss}), the size of $k-$nearest neighborhoods in Eq. (\ref{eq:spreadout}) and the fine cluster number $R$ in Eq. (\ref{eq:refined-similarity}). When evaluating one of the four parameters, we fix the others. 
We evaluated these parameters on the dataset Market1501 and DukeMTMC and compare the performance with mAP. 
We set the parameters $\alpha=0.5$, $\mu=0.1$, $k=6$, $R=5$ on Market1501 and $R=2$ on DukeMTMC-ReID based on the following analysis.

\subsubsection{Loss weight $\alpha$}
\label{sec:param-alpha}
This parameter controls the weight between metric losses under the two kinds of different pseudo label supervision.
As shown in Fig. \ref{fig:param-analysis} (a), 
we can see that when $\alpha$ increases from 0 to 1, the overall performance shows an upward trend. 
When $\alpha=0$, it means only using coarse pseudo labels; When $\alpha=1$, it means only using fine pseudo labels. The performance of using fine pseudo labels outperforms that of only using coarse pseudo labels by a large margin, which shows the effectiveness of our proposed hierarchical clustering mechanism. The performance significantly increases as $\alpha$ increases from 0 to 0.5, and tends to be stable as $\alpha$ increases from 0.5 to 1.
When $\alpha$ increases from 0.5 to 0.1, there are slight fluctuations in performance. We speculate the reasons as follows:
(1) The number of prototypes within each coarse cluster is fixed in all our experiments. If there are huge intra-cluster variances within several noisy hard clusters, the number of the selected prototypes is too small to represent the global characteristics of these noisy clusters. As a result, the lack of more representative prototypes within clusters of huge variances will bring about extra noise.
(2) The slight fluctuations in performance appear when $\alpha$ is higher than 0.5, because the refined pseudo labels will dominate the learning of metric losses if $\alpha>0.5$. Especially in the early training stage, there may be more hard noisy clusters of huge variances. The lack of more robust prototypes within those noisy clusters will degenerate the network's discriminability a little.

\subsubsection{Loss weight $\mu$} As shown in Fig. \ref{fig:param-analysis} (b), 
we evaluate the performance with different values of the parameter $\mu$.
When $\mu=0$ it means learning without the IM-spread-out regularization, and our method gets low performance. As $\mu$ increases, the performance is enhanced greatly and it achieves the peak when $\mu=0.1$. It shows that enforcing the spread-out property on the entire dataset with our specially designed IM-spread-out regularization can help to alleviate the effects of the noisy supervision signal
When $\mu$ is larger than 0.1, the performance will drop consistently, it can be explained that the features are spreading-out too much, which will break the inherent similarities between samples.

\subsubsection{The number $k$ of $k-$nearest neighborhoods} The parameter $k$ is the number of positives which we assume as the $k-$nearest neighborhoods of a sample. 
As shown in Fig. \ref{fig:param-analysis} (c), when $k$ varies from 1 to 10, its performance stays relatively high, however when $k$ goes larger than 10, its performance drops. It shows that the IM-spread-out regularization is robust to the small $k$ values but when $k$ gets larger, it may degenerate the performance because too many neighborhoods means too many noisy positives. 
It should be noted that when $k=0$, our IM-spread-out regularization (Eq. (\ref{eq:spreadout})) degrades into a loss only considering the instance discrimination, \textit{i.e.,} $\mathcal{K}_{i}\leftarrow \{i\}$. Though instance discrimination can still ensure the spread-out property, it will lose the positive-centered property and break the class consistency. Specifically, positive-centered property means that samples of the same label should be concentrated in the feature space, and class consistency means that neighbors locating near the sample tend to have the same label. As a result, only using instance discrimination in UDA re-ID can not enhance the target domain features' discriminability.

\subsubsection{The fine cluster number $R$} 
\begin{figure} [htp]
  \centering
  \includegraphics[width=\linewidth]{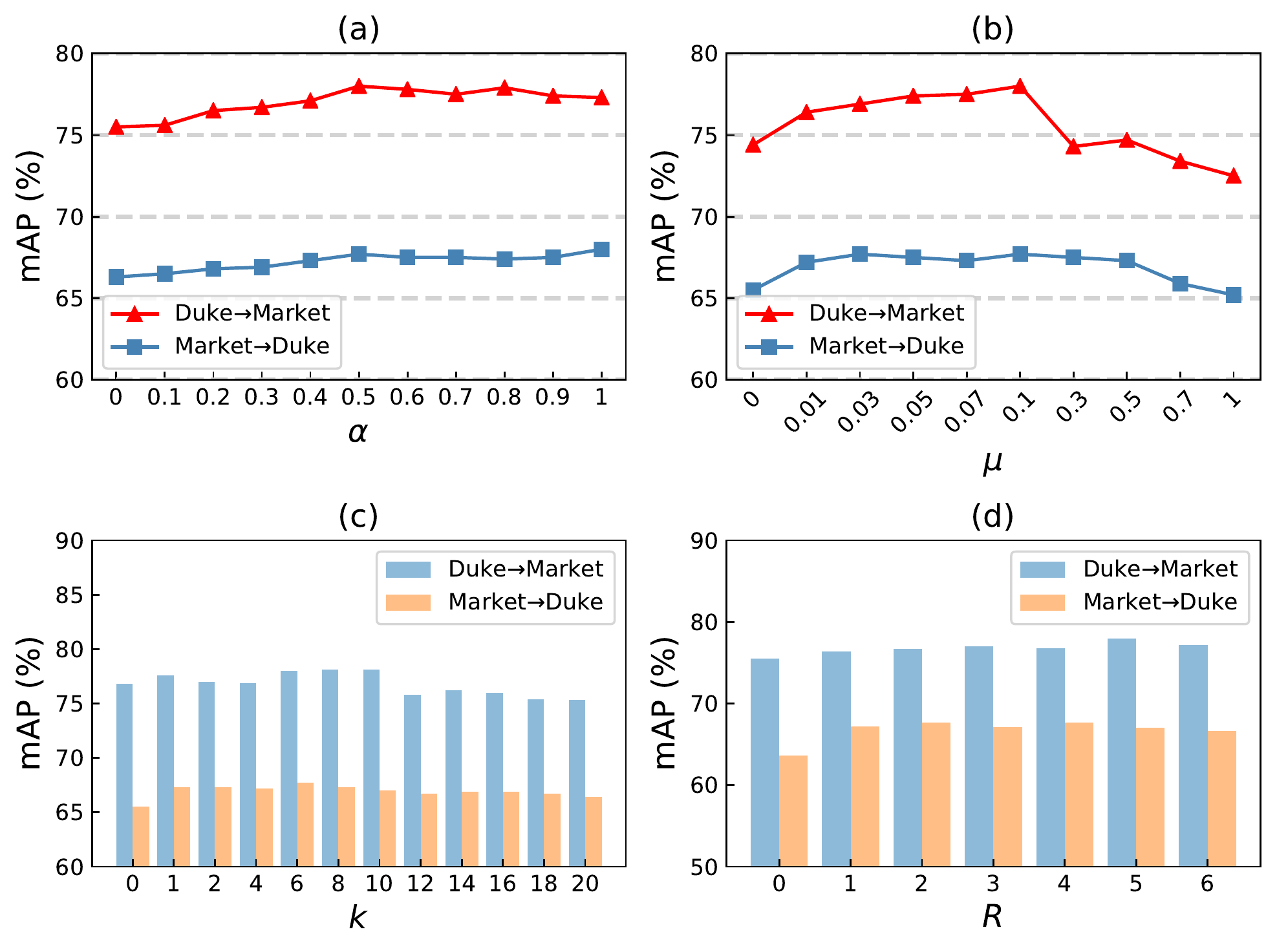}
  \caption{(a) Evaluation on different values of the parameter $\alpha$ in Eq. (\ref{eq:overall-loss}).  (b) Evaluation on different values of the parameter $\mu$ in Eq. (\ref{eq:overall-loss}). (c) Evaluation on different values of $k-$nearest neighborhoods. (d) Evaluation on different fine clustering number $R$.}
  \label{fig:param-analysis}
\end{figure}

Fig. \ref{fig:param-analysis} (d) shows that our method achieves the best performance when $R=5$ for Duke$\to$Marke1501 and $R=2$ for Market1501$\to$Duke. These results can reveal that through hierarchical clustering, we can utilize the hierarchical information from the target data itself to further assign more reliable pseudo labels for samples. 
$R=0$ means only using the one-stage coarse clustering instead of using the two-stage hierarchical clustering, \textit{i.e.}, not performing fine clustering.
$R=1$ means that we calculate the average feature within a coarse cluster and represent it as the fine cluster centroid. 

\section{Conclusion}
In this work, we propose  a  novel  approach  called Dual-Refinement to alleviate  the  pseudo  label  noise in clustering-based UDA re-ID, including the off-line pseudo label refinement to  assign  more  accurate labels and the on-line feature refinement  to alleviate  the  effects  of  noisy  supervision signal.
Specially, we design an off-line pseudo label refining strategy by utilizing the hierarchical information in target domain data. We also propose an on-line IM-spread-out regularization to alleviate the effects of the noisy samples. The IM-spread-out regularization is equipped with an instant memory bank that can consider the entire target data during training. Compared to state-of-the-art methods on UDA re-ID, Dual-Refinement is trained with only a single model and has shown significant improvement of performances.

\bibliographystyle{IEEEtran}

\bibliography{reference}

\end{document}